\documentclass[11pt]{article}
\usepackage{amsmath,amssymb,graphicx}
\usepackage{algorithm}
\usepackage{algorithmic}
\usepackage{subfigure}
\usepackage{enumitem}
\usepackage{color}
\usepackage{multirow}
\usepackage[margin=0.8in]{geometry}
\usepackage{setspace}
\graphicspath{{./Figs/}}

\newcommand{\y}{{\boldsymbol y}}
\newcommand{\x}{{\boldsymbol x}}
\newcommand{\z}{{\boldsymbol z}}
\newcommand{\n}{{\boldsymbol n}}

\newcommand{\setF}{{\mathcal F}}
\newcommand{\setA}{{\mathcal A}}
\newcommand{\setP}{{\mathcal P}}
\newcommand{\setM}{{\mathcal M}}
\newcommand{\R}{{\mathbb R}}
\title{Deep Ptych: Subsampled Fourier Ptychography using Generative Priors}%}via Generative Prior}
\author{Fahad Shamshad, Farwa Abbas, Ali Ahmed \\ \small{Department of Electrical Engineering, Information Technology University, Lahore, Pakistan.} \\ \tt{\small{$\lbrace$fahad.shamshad, farwa.abbas, ali.ahmed$\rbrace$@itu.edu.pk }} }

\begin{document}
%\ninept
%
\maketitle
\begin{abstract}
This paper proposes a novel framework to regularize the highly ill-posed and non-linear Fourier ptychography problem using generative models. We demonstrate experimentally  that our proposed  algorithm, \textit{Deep Ptych}, outperforms the existing Fourier ptychography techniques, in terms of quality of reconstruction and robustness against noise, using far fewer samples. We further modify the proposed approach to allow the generative model to explore solutions outside the range, leading to improved performance.% on rich image datasets not well learned by the generative networks.
\end{abstract}
%
%\begin{keywords}
%Fourier ptychography, phase retrieval, generative models, noise robustness, subsampling
%\end{keywords}
%
\section{Introduction} \label{sec:intro}
%{\let\thefootnote\relax\footnote{$^{\ast} $Equal contribution.}}
Resolution loss in long distance imaging can primarily be attributed to the diffraction blur, that is caused by limited aperture of the imaging system \cite{holloway2017savi}. To mitigate the effects of the diffraction blur, recently an emerging computational imaging technique known as Fourier Ptychography (FP) has shown promising results \cite{holloway2016toward,zheng2013wide}. The objective of FP is to recover a high-resolution image from multiple diffraction-limited low-resolution images. In this paper, we consider recovering the signal $\x \in \mathbb{C}^n$ captured via forward acquisition model of FP, given by:
\begin{equation} \label{eq:pr}
\y_\ell = \vert \setA_\ell (\x) \vert + \n_\ell, \;\; \text{for} \;\; \ell = 1,2,...,L,
\end{equation}
where $\y_\ell \in \mathbb{R}^n$ is diffraction-limited image corresponding to $\ell^{th}$ camera, $\setA_\ell:\mathbb{C}^n \rightarrow \mathbb{C}^n$ is the linear operator representing the forward acquisition model, and $\n_\ell \in \mathbb{R}^n$ denotes  noise perturbation. For $\ell^{th}$ camera, the linear operator $ \setA_\ell$ has the form  $ \setF^{-1} \setP_\ell {\circ} \setF$, where $\setF$ denotes 2D Fourier transform, $\setP_\ell$ is a pupil mask that acts as a bandpass filter in the Fourier domain, and $\circ$ represents the Hadamard product (details are provided in subsequent sections). Specifically, FP works by iteratively stitching together a sequence of frequency limited low-resolution images $\y_\ell$ in Fourier domain to recover the high-resolution true image $\x$. Since optical sensors can measure only the magnitude of the signal \cite{shechtman2015phase}, phase information is lost during the acquisition process --- making the FP problem highly ill-posed.

Traditional approaches to overcome the ill-posedness of FP are iterative phase retrieval algorithms \cite{yang2011iterative}. Generally, phase retrieval algorithms fall into two categories. The $\textit{first}$ approach is to introduce redundancy into the measurement system, where more measurements are taken than the dimension of true signal usually in the form of oversampled Fourier transform \cite{bendory2017fourier}, short-time Fourier transform \cite{bendory2018non}, structured illuminations \cite{candes2015phase}, etc. The $\textit{second}$ approach is to exploit the structural assumptions on the true signal as a prior, such as sparsity \cite{pauwels2018fienup} or non-negativity \cite{beinert2017non}. These prior based phase retrieval approaches have recently attained much significance for the purpose of reducing number of measurements as the acquisition of measurements is usually expensive and time-consuming, especially at high resolutions.

FP recovery algorithms that exploit redundant information perform well provided the set of low-resolution images $\y_\ell$ have high overlapping frequency bands (typically $60\%$ overlap) in the Fourier domain \cite{qian2014efficient}. The reconstruction quality of these algorithms quickly degrades with the decrease in overlap. This high overlap requirement for faithful reconstruction of the true image $\x$ results in long acquisition times and high computational cost \cite{holloway2016toward}.

To reduce the computational cost, algorithms that exploit prior information about true signal have recently been explored for FP. Among them, priors learned from large datasets by utilizing the power of deep neural networks have shown promising results for low overlap case, thus reducing computational cost \cite{kappeler2017ptychnet,nguyen2018deep,boominathan2018phase}. Specifically, these deep learning based approaches invert  the forward acquisition model of FP via end-to-end training of deep neural networks in a supervised manner \cite{lecun2015deep}. However, even a slight change in the noise level $\n_\ell$ or parameters of FP measurement model, such as aperture diameter or overlap would require costly retraining of these models.
%{\color{red}However these neural network based approaches do not incorporate any subsampling strategies for reducing measurements.}
Apart from reducing  overlap, an alternative approach to reduce computational cost for FP, is to design realistic subsampling strategies by exploiting some prior information about the true signal. This alternative has not yet been investigated by the deep learning based FP approaches. Recently, Jagatap et al. \cite{jagatap2018sub} %and Chen et al. \cite{chen2018low} 
explored this alternative by leveraging sparsity prior. The resulting algorithm (CoPRAM - {\textit{Compressive Phase Retrieval using Alternative Minimization}) significantly reduces the number of samples required for faithful reconstruction of the true signal $\x$ in FP setup. However, it has been shown in \cite{metzler2016bm3d} that  sparsity priors often fail to capture the complex structure that many natural signals exhibit resulting in unrealistic signals also fitting the sparse prior modelling assumption.

Recently, neural network based implicit generative models such as Generative Adversarial Networks (GANs) \cite{goodfellow2014generative} and Variational Autoencoders (VAEs) \cite{kingma2013auto} have been quite successful in modelling complex data distributions especially that of images. In these models, the generative part $(G)$, learns a mapping from low dimensional latent space $\z \in \mathbb{R}^{k}$ to a high dimensional sample space $G(\z) \in \mathbb{R}^n$ where $k \ll n$. During training, these generative models are encouraged to produce samples that resemble with that of training data  $\mathcal{X}$. A well-trained generator, given by deterministic function $G: \mathbb{R}^k\rightarrow\mathbb{R}^{n}$  with a distribution $P_Z$ over $\z$ (usually random normal or uniform), is therefore capable of generating fake data indistinguishable from the real data it has been trained on. Due to their power to effectively model natural image distributions, generative priors have recently been introduced to solve ill-posed inverse problems like %denoising\cite{heckel2018deep},
 compressed sensing \cite{bora2017compressed}, phase retrieval \cite{shamshad2018robust, paul2018pg}, blind deconvolution \cite{asim2018solving}, etc. Notably, these generative prior based approaches, have been shown to improve over sparsity-based approaches, thus advancing the state of the art in several fields \cite{hand2017global}.

\begin{figure*}[t] \label{fig:block_dia}
\centering
{\includegraphics[width=1.03\textwidth]{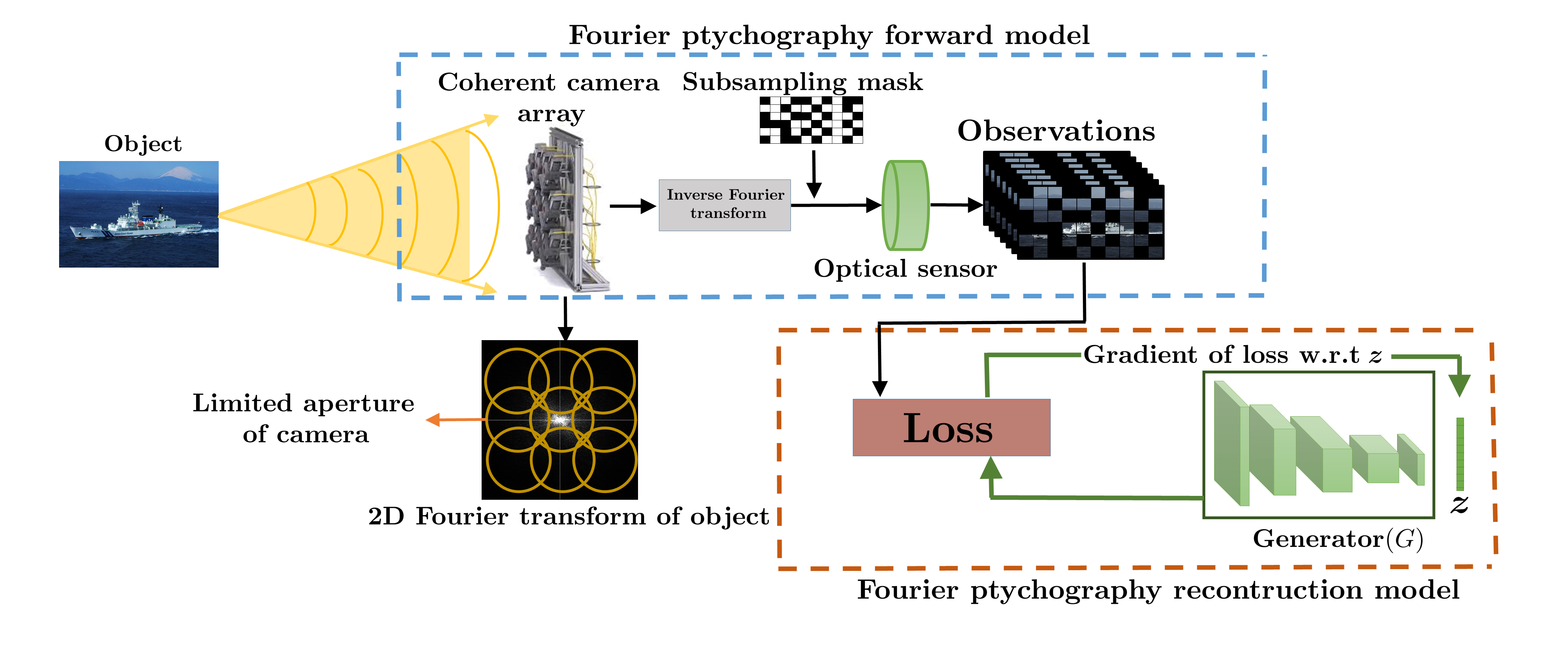}}
\caption{\textit{Overview of Fourier ptychography forward model and proposed reconstruction algorithm. In the acquisition phase, a coherent camera array captures illumination field from the object. Each camera in the array has a limited aperture that captures a portion of 2D Fourier transform of the field. This bandlimited signal is then focused to an image plane where this propagation to the image plane can be modelled as an inverse Fourier transform. After  that, a subsampling operator is applied, and the signal is subsequently captured by an optical sensor that measures its magnitude while discarding the phase. During the reconstruction phase of Fourier ptychography, the generator ($G$) optimizes latent code $\z$ via gradient descent algorithm to find a corresponding $G(\boldsymbol{z})$ that best explains the observations.}}
\end{figure*}
\subsection{Our Contributions}
In this paper, we leverage the expressive power of generative models to regularize FP problem. To the best of our knowledge, this is the first work that explores the generative models for FP setup. In particular, we make the following contributions:
\begin{enumerate}[leftmargin=12pt,nolistsep]
\item We combine the expressive power of deep generative models (GANs and VAEs) with non-linear inverse problem of FP for the first time. We show that the resultant problem can be solved effectively using conventional gradient descent algorithm yielding promising results. Moreover, unlike deep learning based approaches \cite{kappeler2017ptychnet,boominathan2018phase}, our approach does not require expensive retraining in case of slight changes in parameters such as subsampling ratio or noise level.
\item We show experimentally that the proposed approach produces impressive results for low subsampling ratios and high noise. 
%    \item We show that using generative maps induces a very strong prior that is highly robust to noise.
\item Finally, to further improve the performance of the proposed scheme, especially for the rich image datasets not well learned by the generator, we present a modification in the loss function that allows more flexible reconstructions. 
\end{enumerate}

\section{Formulation} \label{sec:formulation}
In this section, we describe the forward acquisition model of FP for long-distance imaging following the setup of \cite{holloway2016toward}, as shown in Fig. \ref{fig:block_dia}. In long-distance imaging, resolution of the image is limited by diffraction of light at the limited aperture of the imaging system. To emulate capturing the scene with a large synthetic aperture, multiple cameras ($L$) are deployed, usually in a square grid. Such an arrangement is called a coherent camera array. In long-distance imaging, coherent camera array is usually placed in far field of the object ($\x$), and the Fraunhofer approximation is satisfied. Under this approximation, illumination field emerging from the object when intercepted by the camera array can be treated as 2D Fourier transform ($\setF$) of the image at the object plane \cite{born2013principles}. Each camera in the array have a limited aperture denoted by $\setP_\ell$ ($\ell = 1,2,...,L$)  that acts as a bandpass filter covering different parts of Fourier domain image as shown in Fig. \ref{fig:block_dia}. The bandlimited signal is then focused to an image plane where this propagation to the image plane can be modelled as an inverse Fourier transform ($\setF^{-1}$). After that, a subsampling operator ($\setM_{\ell}(\cdot)$), is applied to effectively reduce the number of measurements. Subsequently, the complex spatial domain image is captured by an optical sensor that measures only the magnitude while discarding the phase information. Mathematically, for a total of $L$ cameras, observations $\y_\ell$ can be modelled as%are given by {\color{blue}the} model {\color{blue}as shown below}
\begin{equation} \label{eq:forward-model}
\y_\ell = \sum_{\ell=1}^{L} \vert \mathcal{A}_\ell (\x)\vert + \n_\ell,
\end{equation}
where $\setA_\ell = \setM_\ell \setF^{-1} \setP_\ell \circ \setF$ is the measurement model prior to optical sensor acquisition step, $\circ$ denotes the Hadamard product and $\setM_\ell$ is the subsampling operator. Subsampling operator when applied to an image, randomly picks a fraction of samples $(f)$ discarding the others \cite{jagatap2018sub}. We define the subsampling ratio as the fraction of samples retained by $\setM_\ell$ divided by the total number of observed samples i.e. 
\begin{equation}
\text{Subsampling Ratio (\%)} =\frac{\text{Fraction of samples retained ($f$)}\times 100}{\text{Total observed samples ($nL$)}}. \nonumber
\end{equation}

\section{Deep Ptych} \label{sec:deepptych}

In this section, we formally introduce our proposed approach $\textit{Deep Ptych}$. We assume that the true image $\boldsymbol{x} \in \R^n$  is a member of some structured class of images, denoted by $\mathcal{X}$. For example, $\mathcal{X}$ may be set of face images or digits. A generative model is trained on a representative sample set from class $\mathcal{X}$. Given low-dimensional input vector $\boldsymbol{z} \in \mathbb{R}^k$, the generator $G$ after training, generates new samples $G(\boldsymbol{z})$ similar to the representative samples of class $\mathcal{X}$. Once trained, the weights of the generator are kept fixed (pretrained generator). To recover the estimate of true image $\hat{\x}$ from FP measurements $\boldsymbol{y}_\ell$ for $\ell = 1,2,...,L,$ in (\ref{eq:forward-model}), we propose minimizing the following objective function 
\begin{align}\label{eq:Optimization-Ambient}
{\hat{\x}} :=  \underset{\substack{\x \in \text{Range}(G)}}{\text{argmin}} \sum_{\ell=1}^{L} \Vert \y_\ell - \vert \mathcal{A_\ell} (x) \vert \Vert^{2}_2 ,
\end{align}
where Range($G$) is the set of all the images that can be generated by pretrained $G$. In other words, we want to find an image $\hat{\x}$ that is closest to the observed measurements and lies within the range of the generator. Ideally, the range of the generator must comprise of only the samples drawn from the distribution of the images it has been trained on, i.e., $\exists \,\z$ such that $\x = G(\boldsymbol{z})$. It must be noted that constraining the solution ${\boldsymbol{\hat{x}}}$ to lie inside the generator range will force the solution to be a member of image class $\mathcal{X}$ thus, implicitly reducing the solution ambiguities associated with FP \cite{yang2011iterative}.
The minimization program in \eqref{eq:Optimization-Ambient} can be equivalently formulated in the lower dimensional, latent representation $\z$ as follows:
\begin{align}\label{eq:Optimization-latent}
\hat{\z} = \underset{\boldsymbol{z} \in \R^k}{\text{argmin}}
\ 
\sum_{\ell=1}^{L} \Vert \y_\ell - \vert \mathcal{A_\ell} ({G}(\z))\vert \Vert^2_2.
\end{align}

This optimization program modifies the latent representation vector $\boldsymbol{z}$  such that the generator generates an image $G(\z)$ that is consistent with (\ref{eq:Optimization-latent}). Since the latent representation $\z$ is much smaller in dimension as compared to the ambient dimension of the true image $\x$, it not only reduces the number of unknowns in the FP problem but also allows an efficient implementation of gradients using backpropagation  through the generators. Since optimization program in \eqref{eq:Optimization-latent} is non-convex and non-linear owing to the modulus  operator and non-linear generative model, we resort to the gradient descent algorithm to find an optimal $\hat{\z}$.  The estimated image $\hat{\x}$ is acquired by a forward pass of the $\hat{\z}$ through the generator $G$. Mathematically, $\hat{\x} = G(\hat{\z})$. 
The gradient descent scheme, namely, $\textit{Deep Ptych}$ is formally given in Algorithm \ref{alg:AltGradDescent}, where the objective function in \eqref{eq:Optimization-latent} is denoted by $\mathcal{L}(\z)$, that is, $\mathcal{L}(\z) : = \sum_{\ell=1}^{L} \|\y_\ell -| \mathcal{A_\ell} ({G}(\z))| \|^2_2$, and $\nabla \mathcal{L}(\z)$ denotes its gradient w.r.t. $\z$. 

%Since the range of the generative models can be traversed by a much lower dimensional latent representations compared to the ambient dimension of the images, it not only reduces the number of unknowns in the FP problem but also allows for an efficient implementation of gradients in this lower dimensional space using back propagation through the generators.

%Mathematically, $\boldsymbol{x}^* \approx G(\boldsymbol{z}^*)$.

 %Note that due to its non-convex nature \ref{eq:Optimization-latent} can stuck in local minimas instead of reaching true global minima. Different initialization techniques have been proposed to find good initial guess of phase retrieval algorithm that will guarantee convergence. We circumvent this issue by running $L$ gradient descent steps from $R$ different random initialization of $\boldsymbol{z}$ denoted in Figure \ref{fig:proposed_approach1} by set $\mathcal{Z}_0$ where  $\mathcal{Z}_0 = \lbrace \boldsymbol{z}_0^i\rbrace_{i=1}^{R}$. Optimal $\boldsymbol{z}$, denoted as $\boldsymbol{z}^*$, is the one that gives minimum reconstruction error. The desired solution is $\boldsymbol{x}^* = G(\boldsymbol{z^*})$.

\begin{algorithm}[t] \label{alg:AltGradDescent}
	\caption{{Deep Ptych}}
	\begin{algorithmic} 
		\STATE \textbf{Input:} $\y$, $\mathcal{A}$, $G$, and $\eta$ \\
		\textbf{Output:}  Estimate $\hat{\x}$  \\
		\STATE \textbf{Initialize:}\\
		$\z_0 :=  \mathcal{N}(0,I_k)$ 
		\FOR{${t = 1,2,3,\ldots T}$}
		\STATE{$ \boldsymbol{z}_{t+1} \leftarrow \boldsymbol{z}_{t}$ - $\eta \nabla_{\z_{t}} \mathcal{L}(\boldsymbol{z}_{t})$. } %\quad (\ref{eq:Optimization-latent2})
		\ENDFOR \\
		$\hat{\x}  \leftarrow G(\z_T)$
   \end{algorithmic}
   \label{alg:AltGradDescent}
\end{algorithm}

\subsection{Beyond the range of generator: \textit{Deep Ptych+}}
The shortcoming of the generative network in effectively learning the distribution of the image dataset, especially more complex ones, hampers its ability to reliably reproduce a given new (test) sample from its range \cite{dhar2018modeling}. To address this shortcoming, we allow the reconstructed image to deviate a bit from the range of the generator if it improves the measurement loss. To achieve this, we propose solving the following modified version of the optimization program in  \eqref{eq:Optimization-latent} 
\begin{equation}\label{eq:slack-optimization}
(\hat{\x},\hat{\z}) = \underset{\z \in \R^k,\,\x \in \R^n}{\text{argmin}} \ \sum_{\ell=1}^{L} \|\y_\ell -| \mathcal{A_\ell} (\x)| \|^2_2 + \lambda \|\x - G(\z)\|_2^2.
\end{equation}
The first term in the objective favors an image $\x$ with smaller measurement loss, while the second term ensures that in doing so such an $\x$ should not deviate too far away from the range of the pretrained generative network. The free parameter $\lambda \geq 0$ controls the degree to which we are willing to deviate away from the range of the generator, for example, one might choose a smaller $\lambda$ when the generator is known to not reliably reproduce the image samples from the dataset. 

To minimize (\ref{eq:slack-optimization}), we use an alternating gradient descent scheme that takes gradient steps in one of the unknowns $\x$ or $\z$ while keeping the other fixed.} This enables us to land at the local minima $(\hat{\x},\hat{\z})$ of the objective in \eqref{eq:slack-optimization}. We take $\hat{\x}$ to be the reconstructed image, which generally improves over the reconstructions from Deep Pytch algorithm above. We refer to this modified approach as Deep Ptych+.

\section{Experimental results}
We now provide simulation experiments to evaluate the performance of $\textit{Deep Ptych}$ and $\textit{Deep Ptych+}$. %under generative priors. 
To quantitatively evaluate the performance of our algorithm, we use two metrics Peak Signal to Noise Ratio (PSNR) and Structural Similarity Index Measure (SSIM). In all our experiments, we report results on a held out test set, unseen by the generative model during training.
\subsection{Datasets and Generative models}
We use three datasets (MNIST \cite{lecun1998gradient}, CelebA (64$\times$64) \cite{liu2015deep}, and high-resolution CelebA (128$\times$128) \cite{karras2017progressive}) and two generative models (VAE and GAN). This provides some evidence that our approach is generalizable to many types of datasets and generative models.

%We demonstrate effectiveness of Deep-Ptych on two bench mark datasets: MNIST and CelebA.
\textbf{MNIST with VAE}: MNIST consists of 28$\times$28 grayscale images of handwritten digits, with 60,000 training and 10,000 test examples. We super-resolve original MNIST dataset to size 56$\times$56 using super-resolution method proposed in \cite{cruz2018single}, that gives promising results without requiring external training data. We train VAE, having same architecture as proposed in \cite{asim2018solving} for image datasets, on this super-resolved dataset with size of latent dimension set to 50, batch size of 128, $\lambda$ set to  $2 \times 10^{-4}$, and learning rate of $5 \times 10^{-4}$ with Adam optimizer. 

\textbf{CelebA with DCGAN}: We use aligned and cropped version of CelebA dataset having more than 200,000 RGB face images of size 64$\times$64$\times$3. Each pixel value is scaled between $[-1,1]$. Deep convolutional generative adversarial network (DCGAN) is trained on this rescaled dataset with architecture as proposed in \cite{radford2015unsupervised}  having latent dimension size set to 100, batch size of 64, $\beta_1$ set to 0.5, $\lambda$ set to  $1.4 \times 10^{-4}$, and learning rate of $2 \times 10^{-4}$ using Adam optimizer.  %DCGAN model is trained by updating generator $G$ twice and discriminator D once in each cycle to avoid fast convergence of $D$. 
%Each update during training use Adam optimizer [56] with batch size 64, $\beta_1$ = 0.5, and learning rate 0.0002.
\begin{figure*}[t] 
\centering
\subfigure[CelebA]{\includegraphics[height = 5.0cm, width = 7.8cm]{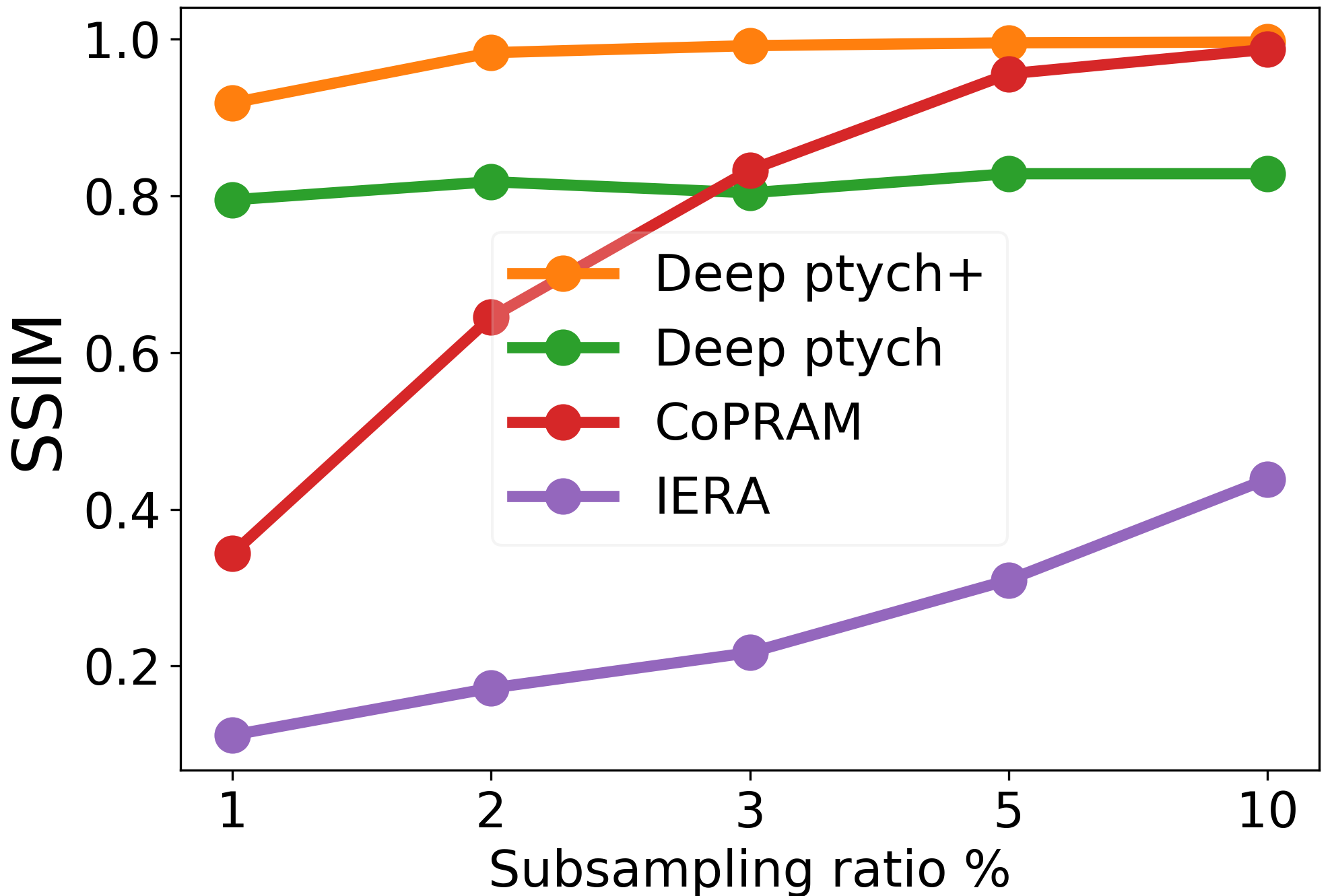}}
\subfigure[MNIST]{\includegraphics[height = 5.0cm, width = 6.7cm]{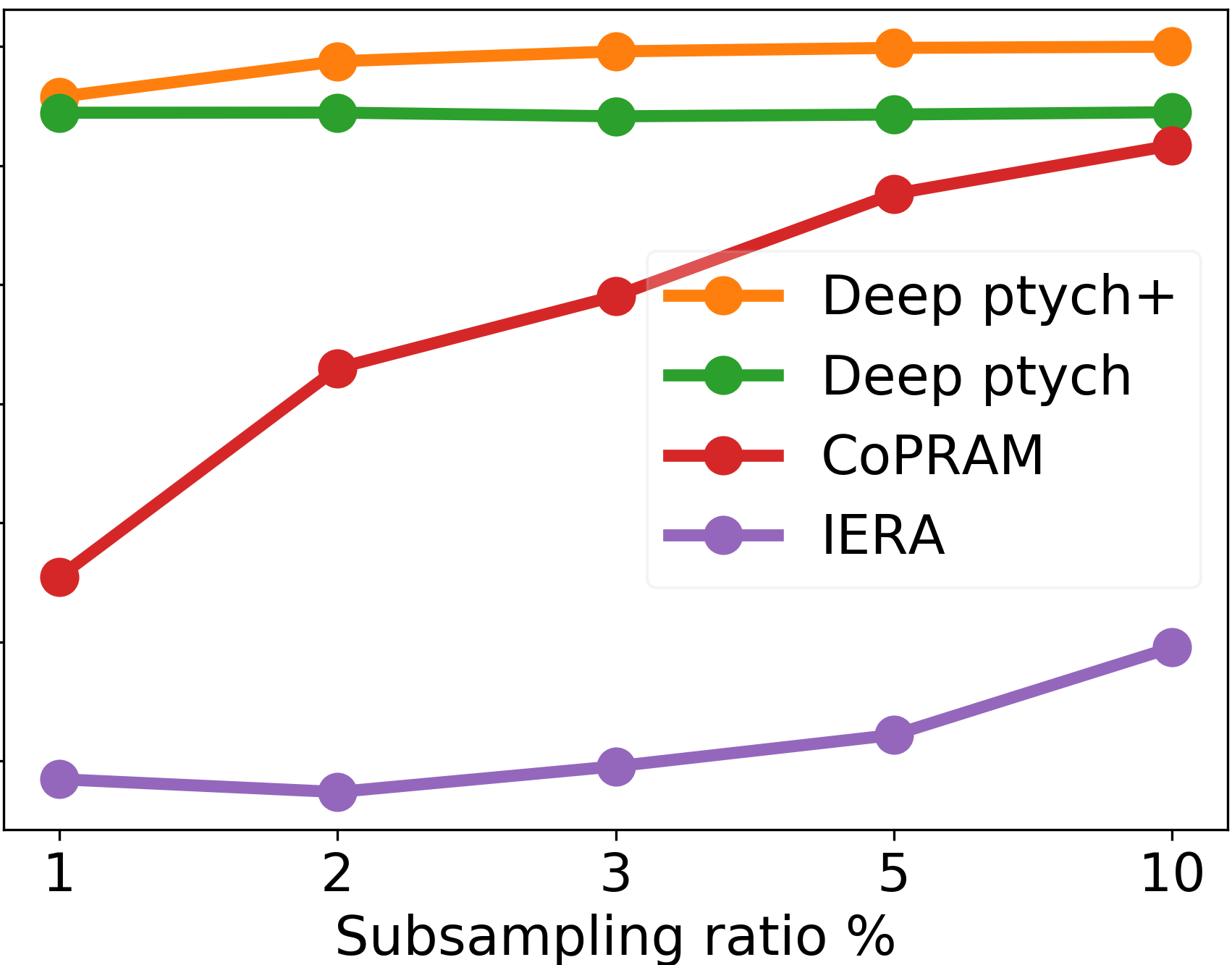}} \\
\subfigure[CelebA]{\includegraphics[height = 5.0cm, width = 7.8cm]{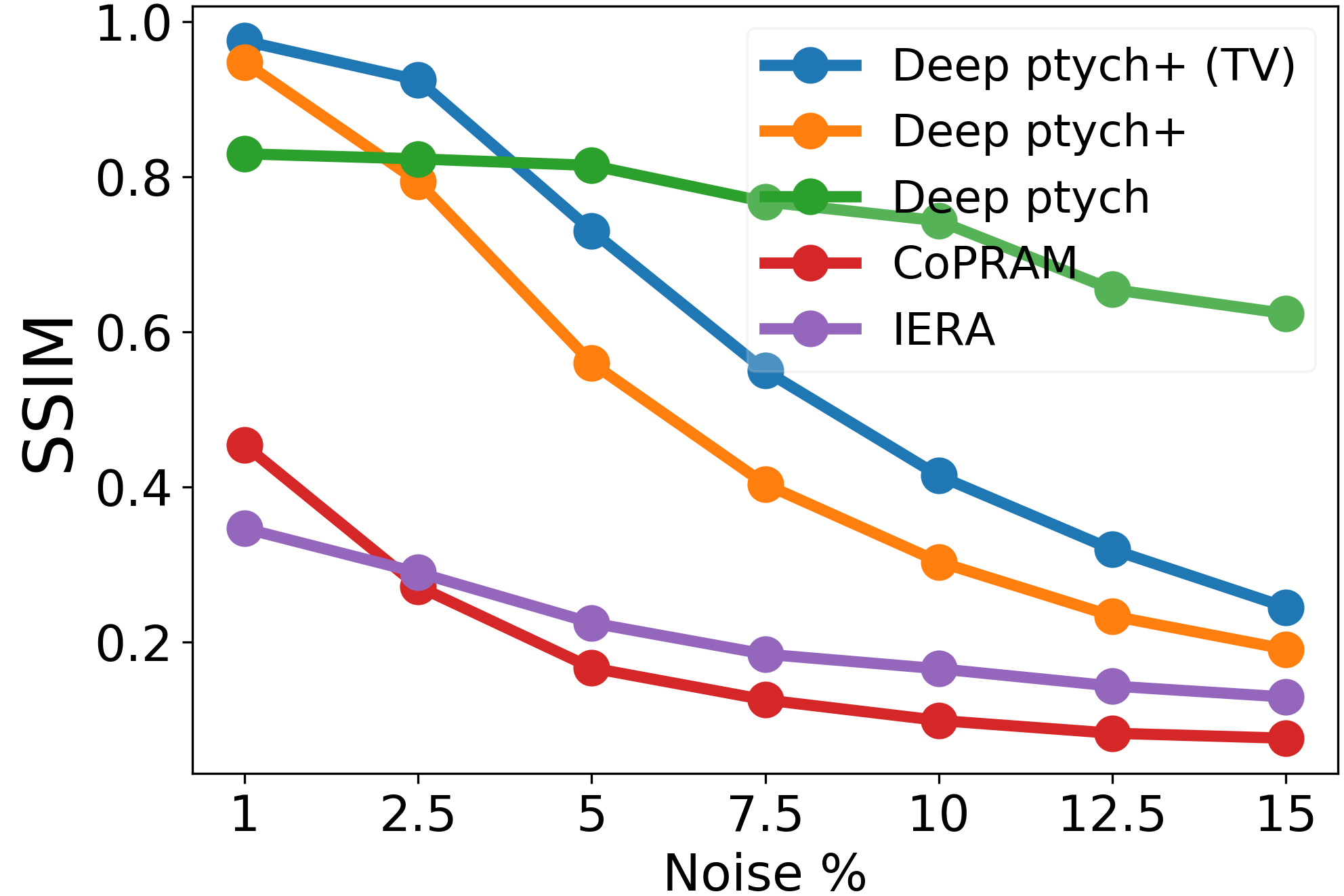}}
\subfigure[MNIST]{\includegraphics[height = 5.0cm, width = 6.7cm]{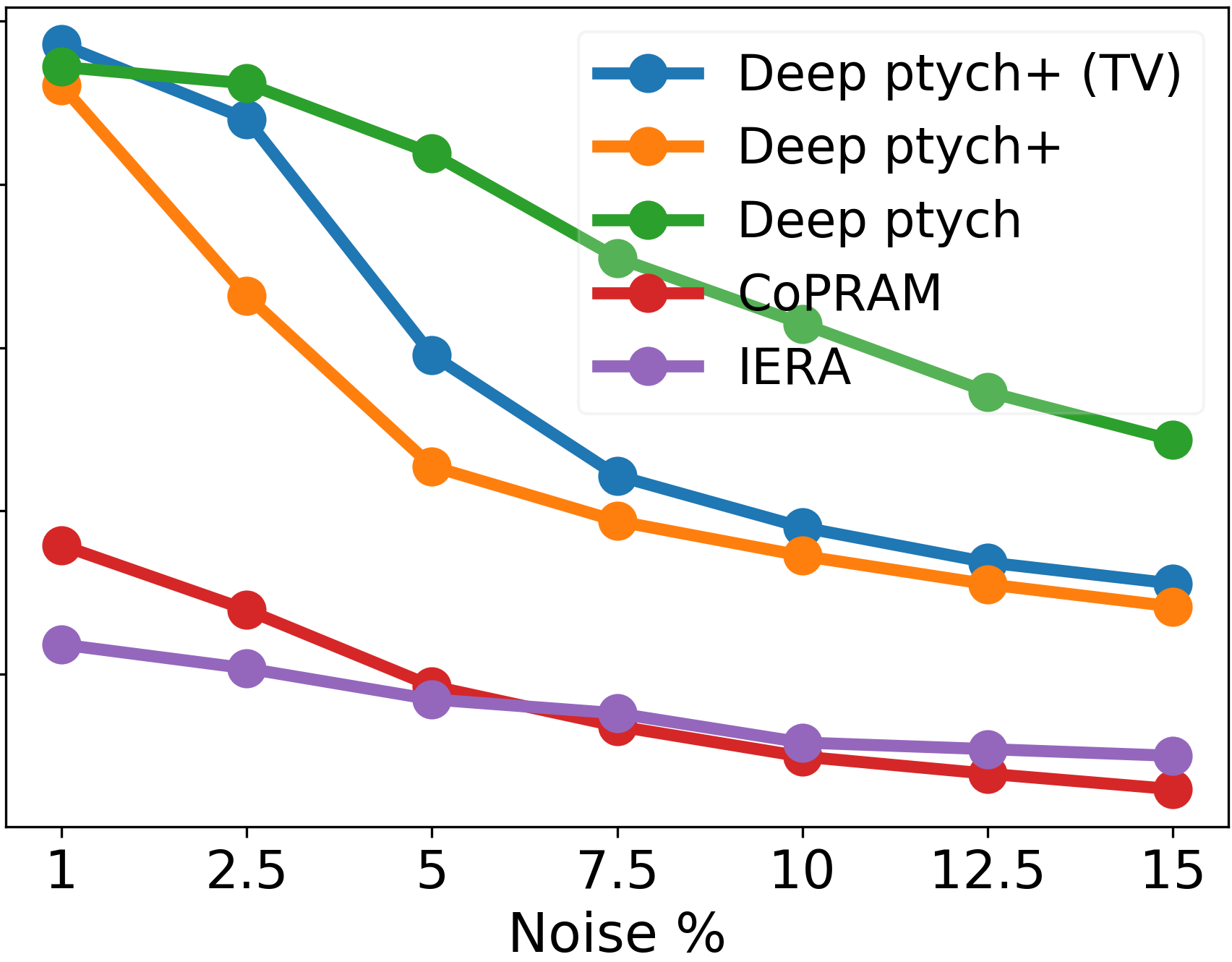}}
%\subfigure[CelebA]{\includegraphics[height = 6.0cm, width = 4.8cm]{celeba_sub.png}}
%\subfigure[MNIST]{\includegraphics[height = 6.0cm, width = 3.6cm]{mnist_sub.png}} 
%\subfigure[CelebA]{\includegraphics[height = 6.0cm, width = 3.8cm]{celeba_noise1.png}}
%\subfigure[MNIST]{\includegraphics[height = 6.0cm, width = 3.6cm]{mnist_noise.png}}
\caption{\textit{SSIM plots of proposed approaches against baseline methods for different subsampling ratios and noise levels  for CelebA ((a) and (c)) and MNIST ((b) and (d)) respectively. Results are averaged over 10 test images of each dataset.}}
\label{fig:ssim_plots}
\end{figure*}

\textbf{CelebA (128$\times$128) with Progressive GAN \cite{karras2017progressive}}: To show the dependence of \textit{Deep Ptych} on the expressive power of generator, we use a powerful generative model named progressive GAN \cite{karras2017progressive}, trained on 128$\times$128 face images. We use the originally proposed architecture having the latent dimension set to 512.
%We use architecture as proposed in original paper \cite{karras2017progressive}, having a latent dimension set to 512. %We will show that this generator effectively eliminate range error.

%\textbf{Generator Architectures} For grayscale datasets, architectures for generator and discriminator are given in Table 1. Size of low-dimensional vector $\z$ is 40 and sampled from random uniform distribution. Adam optimizer has been used for training with learning rate 0.0002, $\beta_1$ = 0.5, batch size 256 and number of epochs 100. For RGB datasets we use deep convolutional generative adversarial network (DCGAN) model of [55]. Size of low dimensional feature representation z is set to 100 and sampled from random normal distribution. DCGAN model is trained by updating generator $G$ twice and discriminator D once in each cycle to avoid fast convergence of $D$. Each update during training used the Adam optimizer [56] with batch size 64, $\beta_1$ = 0.5, and learning rate 0.0002. For all experiments we use $\lambda$ = 0.001 and $\gamma$ = 0.001. 

\subsection{Baseline methods} %In order to compare the performance of proposed methods, 
We consider IERA (Iterative Error Reduction Algorithm) \cite{holloway2016toward} and CoPRAM \cite{jagatap2018sub} as baseline methods. IERA solves FP problem by alternatively enforcing spatial and Fourier domain constraints whereas CoPRAM exploits the underlying sparse structure of true signal for faithful reconstruction at low subsampling ratios. For CoPRAM, we assume sparsity for MNIST and CelebA in spatial and Fourier domains respectively. We use the default algorithmic parameters for IERA and CoPRAM, unless stated otherwise.

%\subsection{Subsampling results}

%\textbf{MNIST} For [] we change the sparsity to 0.15 that we find to work best with MNIST dataset.

%\begin{figure*}[t] 
%\centering
%\subfigure[CelebA]{\includegraphics[height = 3.8cm, width = 4.5cm]{celeba_sub.png}}
%\subfigure[MNIST]{\includegraphics[height = 3.8cm, width = 4.5cm]{mnist_sub.png}}
%\subfigure[CelebA]{\includegraphics[height = 3.8cm, width = 4.5cm]{celeba_noise.png}}
%\subfigure[MNIST]{\includegraphics[height = 3.8cm, width = 4.5cm]{mnist_noise.png}}
%\caption{\textit{SSIM plots of proposed approaches against baseline methods for different subsampling ratios and %noise levels  for CelebA ((a) and (c)) and MNIST ((b) and (d)) respectively. Results are averaged over 10 test %images of each dataset.}}
%\label{fig:ssim_plots}
%\end{figure*}

\begin{figure*}[h]
\centering
%\raisebox{0.2in}{\rotatebox[origin=t]{90}{\small Subsampling}} \vspace{0em}
\subfigure{\includegraphics[width = 0.17\columnwidth]{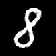}}
\subfigure{\includegraphics[width = 0.17\columnwidth]{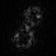}}
\subfigure{\includegraphics[width = 0.17\columnwidth]{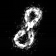}}
\subfigure{\includegraphics[width = 0.17\columnwidth]{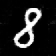}}
\subfigure{\includegraphics[width = 0.17\columnwidth]{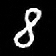}} \\[-0.25em]

\setcounter{subfigure}{0}
\subfigure[\scriptsize Original]{\includegraphics[width = 0.17\columnwidth]{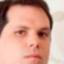}}
\subfigure[\scriptsize IERA]{\includegraphics[width = 0.17\columnwidth]{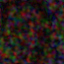}}
\subfigure[\scriptsize CoPRAM]{\includegraphics[width = 0.17\columnwidth]{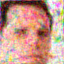}}
\subfigure[\scriptsize DP]{\includegraphics[width = 0.17\columnwidth]{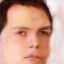}}
\subfigure[\scriptsize DP+]{\includegraphics[width = 0.17\columnwidth]{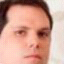}} \\

\subfigure{\includegraphics[width = 0.15\columnwidth]{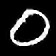}}
\subfigure{\includegraphics[width = 0.15\columnwidth]{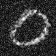}}
\subfigure{\includegraphics[width = 0.15\columnwidth]{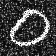}}
\subfigure{\includegraphics[width = 0.15\columnwidth]{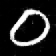}}
\subfigure{\includegraphics[width = 0.15\columnwidth]{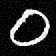}}
\subfigure{\includegraphics[width = 0.15\columnwidth]{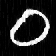}} \\[-0.4em]

\setcounter{subfigure}{0}
\subfigure[\scriptsize Original]{\includegraphics[width = 0.15\columnwidth]{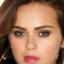}}
\subfigure[\scriptsize IERA]{\includegraphics[width = 0.15\columnwidth]{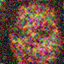}}
\subfigure[{\scriptsize CoPRAM}]{\includegraphics[width = 0.15\columnwidth]{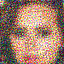}}
\subfigure[\scriptsize DP]{\includegraphics[width = 0.15\columnwidth]{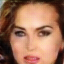}}
\subfigure[\scriptsize DP+]{\includegraphics[width = 0.15\columnwidth]{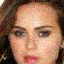}}
\subfigure[\scriptsize DP+(TV)]{\includegraphics[width = 0.15\columnwidth]{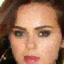}}

\caption{\textit{ (\textbf{Top}): Subsampling results for MNIST and CelebA for $2\%$ subsampling ratio. Results of Deep Ptych (DP)(d) and Deep Ptych+ (DP+)(e) are far superior as compared to IERA (b) and CoPRAM (c). (\textbf{Bottom}): Noise results for MNIST and CelebA for $2.5\%$ noise. Results of DP, DP+, and DP+ (Total Variation) as shown in (d), (e), and (f) are visually appealing as compared to IERA (b) and CoPRAM (c).}}
\label{fig:res}
\end{figure*}

\begin{figure}[h] 
\centering
\subfigure[Original]{\includegraphics[height = 3.95cm, width = 3.95cm]{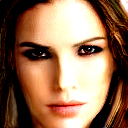}}
\subfigure[IERA]{\includegraphics[height = 3.95cm, width = 3.95cm]{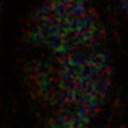}}
\subfigure[CoPRAM]{\includegraphics[height = 3.95cm, width = 3.95cm]{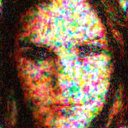}}
\subfigure[Deep Ptych]{\includegraphics[height = 3.95cm, width = 3.95cm]{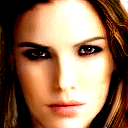}}
\caption{\textit{Deep Ptych results for $128 \times 128$ face images with progressive GAN generator for $1\%$ subsampling ratio and aperture diameter of 30 pixels. Using powerful generator, results are realistic and visually appealing as compared to IERA (b) and CoPRAM (c).}}
\label{fig:pgan}
\end{figure}

%\subsection{Results on CelebA} 
%It can be noted that we can also impose sparsity in a wavelet basis (such as Haar) and we expect to achieve similar improvements in the SSIM.

\subsection{Subsampling results}
We observed that CoPRAM does not work well with additive noise therefore, to carry out a fair comparison, we perform all subsampling experiments without noise.
%Our experiments menifest that even without noise, \textit{Deep ptych} and \textit{Deep ptych+} outperform IERA and CoPRAM at low {\color{red}undersampling ratio.}
For MNIST, we run gradient descent algorithm using Adam optimizer with learning rate of 0.05 for 2000 steps. We set aperture diameter to 15 pixels and overlap to 65$\%$ with camera array consisting of $81$ cameras (square grid of $9\times9$). We use same parameters for CelebA, except learning rate and aperture diameter, that are set to 0.01 and 16 pixels respectively.

%For MNIST and CelebA, quantitative results in terms of SSIM and PSNR for different subsampling rates  are shown in Fig. \ref{fig:ssim_plots} and Table \ref{tab:psnr_plots} respectively.
We observe that \textit{Deep Ptych} and \textit{Deep Ptych+} are able to attain much higher SSIM and PSNR values at low subsampling ratios as shown in Fig. \ref{fig:ssim_plots} and Table \ref{tab:psnr_plots} respectively. Qualitative results at $2\%$ subsampling ratio show faithful reconstruction of \textit{Deep Ptych} and \textit{Deep Ptych+} as compared to IERA and CoPRAM results.
%, that are blurry and full of artifacts. 
Since output of \textit{Deep Ptych} is constrained to lie in generator range, increasing subsampling ratio results in its performance saturation. CoPRAM has no such limitation so its performance eventually surpasses $\textit{Deep ptych}$ at higher subsampling ratios. We show in Fig. \ref{fig:pgan} that for a more expressive generator this range constraint becomes less effective, resulting in visually appealing reconstructions for \textit{Deep Ptych}. \textit{Deep Ptych+}, on the other hand, outperforms IERA and CoPRAM at all subsampling ratios.%  as it leverages the extra information at high sampling rate and at same time exploiting rich structure imposed by generative priors.}

\begin{table}[h]
\centering
\scalebox{1}{
\begin{tabular}{|c|c|c|c|c|c|c|c|c|}
\hline
\multirow{3}{*}{\textbf{}} & \multicolumn{4}{c|}{\textbf{MNIST}}                                             & \multicolumn{4}{c|}{\textbf{CelebA}}                                            \\ \cline{2-9} 
                           & \multicolumn{2}{c|}{\textbf{Subsampling}} & \multicolumn{2}{c|}{\textbf{Noise}} & \multicolumn{2}{c|}{\textbf{Subsampling}} & \multicolumn{2}{c|}{\textbf{Noise}} \\ \cline{2-9} 
                           & \textbf{1\%}        & \textbf{3\%}        & \textbf{1\%}     & \textbf{10\%}    & \textbf{1\%}        & \textbf{3\%}        & \textbf{1\%}     & \textbf{10\%}    \\ \hline
\textbf{IERA}              & 10.59               & 11.72               & 14.15            & 9.59             & 5.89                & 7.29                & 12.36            & 11.25            \\ \hline
\textbf{CoPRAM}            & 14.17               & 19.16               & 16.63            & 6.45             & 13.95               & 23.85               & 17.53            & 9.34             \\ \hline
\textbf{DP}                & 24.70               & 24.80               & 25.31            & \textbf{19.27}   & 22.54               & 22.79               & 23.57            & \textbf{21.21}   \\ \hline
\textbf{DP+}               & \textbf{29.78}      & \textbf{39.39}      & 34.83   & 15.20            & \textbf{27.99}      & \textbf{38.72}      & 32.08            & 14.00            \\ \hline
\textbf{DP+(TV)}           & -                   & -                   & \textbf{35.98}             & 16.82            & -                   & -                   & \textbf{33.61}   & 16.46            \\ \hline
\end{tabular}
}
\caption{{PSNR (dB) results for MNIST and CelebA.}}
\label{tab:psnr_plots}
\end{table}
\subsection{Noise results}
%We evaluate noise robustness of proposed approaches with varying noise levels. 
For all experiments in this section a noise of $1\%$, for image scaled between 0 to 1, translates to Gaussian noise with zero mean and a standard deviation of $0.01$. All parameter values are set to same as in subsampling experiments with subsampling ratio set to $10\%$. Quantitative results in Fig. \ref{fig:ssim_plots} and Table \ref{tab:psnr_plots} manifest that 
%due to its ability to constraining solution in generator range, 
\textit{Deep Ptych} exhibits far superior performance than baseline approaches for high noise levels. On the other hand, performance of \textit{Deep Ptych+} is superior to IERA and CoPRAM, but degrades quickly by increasing noise percentage. This is because its output is not constrained to lie in generator range (that spans the set of representative samples of noiseless training data). We find that by adding total variation (TV) regularization  (with weighting factor of $10^{-4}$) to \textit{Deep Ptych+} loss function, the performance is improved (generally $0.5-2$dB gain in PSNR) at all noise levels. We dubbed this modified \textit{Deep Ptych+} as \textit{Deep Ptych+ (TV)}. Qualitatively, as depicted in Fig. \ref{fig:res}, the proposed approaches outperform baseline methods in terms of quality of reconstruction at $2.5\%$ noise.%, show superior performance of proposed approaches. %\textit{Deep Ptych} results are sharper but constrained to lie in generator range. 
%Noise artifacts present in \textit{Deep Ptych+} results are removed by adding TV regularization. 

%\section{Conclusion}
%To conclude, we integrate deep generative models to solve Fourier ptychography problem. We experimentally show effectiveness of proposed 

\section{Conclusion}
To conclude, we demonstrated the effectiveness of integrating deep generative priors with Fourier ptychography problem. We showed that the proposed approach is effective at low subsampling ratios and is highly robust to noise. We further refined the proposed algorithm to allow the generative model to explore solutions outside the range, leading to improved performance.%{\color{cyan}Furthermore, the proposed method does not require costly retraining for different measurement models and varying noise levels, unlike existing deep learning based models that are trained in an end-to-end manner.}
\section*{Acknowledgement}
We gratefully acknowledge the support of the NVIDIA Corporation for the donation of NVIDIA TITAN Xp GPU for this research.
\bibliographystyle{IEEEbib}
\bibliography{refs}

\end{document}